%% file: main.tex
\documentclass[10pt,twocolumn,letterpaper]{article}

\usepackage{iccv}
\usepackage{times}
\usepackage{epsfig}
\usepackage{graphicx}
\usepackage{amsmath}
\usepackage{amssymb}
\usepackage{color}
\usepackage{makecell}
\usepackage{multirow}
\usepackage{multicol}
\usepackage{pifont}       % \ding{xx}
\usepackage{bbding}       % \Checkmark,\XSolid,... (需要和pifont宏包共同使用)
\usepackage{fontawesome}  % \faCheck,\faTimes
\usepackage{colortbl}  %彩色表格需要加载的宏包
\usepackage[table]{xcolor}
\usepackage{float}
\usepackage{subfig}
\usepackage{tabularx}

\newcommand{\tablestyle}[2]{\setlength{\tabcolsep}{#1}\renewcommand{\arraystretch}{#2}\centering\footnotesize}
\definecolor{baselinecolor}{gray}{.9}
\newcommand{\baseline}[1]{\cellcolor{baselinecolor}{#1}}
\newcolumntype{x}[1]{>{\centering\arraybackslash}p{#1pt}}
\newcolumntype{y}[1]{>{\raggedright\arraybackslash}p{#1pt}}
\newcolumntype{z}[1]{>{\raggedleft\arraybackslash}p{#1pt}}

\newcommand{\cmark}{\ding{51}}

\usepackage[pagebackref,breaklinks,colorlinks]{hyperref}

\iccvfinalcopy % *** Uncomment this line for the final submission

\newlength\savewidth\newcommand\shline{\noalign{\global\savewidth\arrayrulewidth
  \global\arrayrulewidth 1pt}\hline\noalign{\global\arrayrulewidth\savewidth}}
\def\iccvPaperID{****} % *** Enter the ICCV Paper ID here

\ificcvfinal\fi

\begin{document}

\input{0_metadata}
\input{0_abstract}
\input{1_intro}

% \clearpage
\input{2_related}
\input{3_method}

\input{4_results}

\input{5_conclusions}

{\small
\bibliographystyle{ieee_fullname}
\bibliography{3D_Multi_Camera,egbib}
}

\end{document}

%% file: 0_metadata.tex
%%%%%%%%% TITLE
\title{Temporal Enhanced Training of Multi-view 3D Object Detector via Historical Object Prediction}

\author{
Zhuofan Zong\textsuperscript{1}\thanks{Equal contribution. \textsuperscript{\dag}Corresponding authors.}
\quad Dongzhi Jiang\textsuperscript{1}\footnotemark[1]
\quad Guanglu Song\textsuperscript{1}
\quad Zeyue Xue\textsuperscript{2}
\\
\quad Jingyong Su\textsuperscript{3}
\quad Hongsheng Li\textsuperscript{4}\textsuperscript{\dag}
\quad Yu Liu\textsuperscript{1}\textsuperscript{\dag}
\\
\textsuperscript{1}{SenseTime Research}
\quad \textsuperscript{2}{The University of Hong Kong} \\
\textsuperscript{3}{Harbin Institute of Technology, Shenzhen} \quad 
\textsuperscript{4}{The Chinese University of Hong Kong} \\
\small{\texttt{\{zongzhuofan,jdzcarr7,liuyuisanai\}@gmail.com}
\quad \texttt{songguanglu@sensetime.com}} \\
\small{\texttt{xuezeyue@connect.hku.hk} 
\quad \texttt{sujingyong@hit.edu.cn}
\quad \texttt{hsli@ee.cuhk.edu.hk}}
}

\maketitle
\ificcvfinal\thispagestyle{empty}\fi

%% file: 0_abstract.tex
\begin{abstract}
In this paper, we propose a new paradigm, named Historical Object Prediction (HoP) for multi-view 3D detection to leverage temporal information more effectively. 
The HoP approach is straightforward: given the current timestamp $t$, we generate a pseudo Bird's-Eye View (BEV) feature of timestamp $t$-$k$ from its adjacent frames and utilize this feature to predict the object set at timestamp $t$-$k$. 
Our approach is motivated by the observation that enforcing the detector to capture both the spatial location and temporal motion of objects occurring at historical timestamps can lead to more accurate BEV feature learning.
First, we elaborately design short-term and long-term temporal decoders, which can generate the pseudo BEV feature for timestamp $t$-$k$ without the involvement of its corresponding camera images. 
Second, an additional object decoder is flexibly attached to predict the object targets using the generated pseudo BEV feature. 
Note that we only perform HoP during training, thus the proposed method does not introduce extra overheads during inference.
As a plug-and-play approach, HoP can be easily incorporated into state-of-the-art BEV detection frameworks, including BEVFormer and BEVDet series. Furthermore, the auxiliary HoP approach is complementary to prevalent temporal modeling methods, leading to significant performance gains.
Extensive experiments are conducted to evaluate the effectiveness of the proposed HoP on the nuScenes dataset.
We choose the representative methods, including BEVFormer and BEVDet4D-Depth to evaluate our method.
Surprisingly, HoP achieves 68.5\% NDS and 62.4\% mAP with ViT-L on nuScenes test, outperforming all the 3D object detectors on the leaderboard. 
Codes will be available at \url{https://github.com/Sense-X/HoP}.
\end{abstract}

% Our HoP approach is simple: given the current timestamp $t$, we generate the BEV feature of timestamp $t$-$k$ by its adjacent frames and then using it to predict the object set in $t$-$k$.
% This is inspired by our observation that forcing the detector to capture the spatial location and temporal motion of objects in the historical timestamp benefits in more accurate BEV feature learning.

%% file: 1_intro.tex
\section{Introduction}
\label{sec:intro}
% + 介绍Multiview BEV检测方法
%   + 基于纯视觉的3D检测任务在自动驾驶的实际应用中很有价值
%   + 近来基于构建BEV的Multiview的检测方法受到了广泛的关注
% + 介绍最近工作的时序利用方式
%   + 最开始对时序的利用比较朴素，是直接将历史帧和当前帧拼起来
%   + 最近的工作尝试使用stereo的方法，这种方法是为了预测出更准确的深度

% + 对比并引出我们的时序利用方式
%   + 我们提出一种增强时序特征学习的辅助任务，缺失帧预测
%   + 缺失帧预测是使用某一帧相邻帧的BEV来预测出这一帧的BEV，并通过预测出来的BEV检测这一帧的BBox。
%   + 这种缺失帧预测的任务的作用是：BEV上对物体的表示必须足够清晰才能让模型感知到某一物体在time window内的motion情况，从而能预测出缺失的那一帧。经过这个附加任务训练的BEV可以更有利于实际检测中的时序融合的步骤。
%   + 与之前的方法不同的是，这一方式并不是为了在推理的时候能从邻帧获取到更好的信息，而是在训练的时候利用已知的邻帧对训练进行辅助。因此，这一任务的引入对于推理过程没有开销，可以即插即用。（这一段和第二个contribution有点不太相容）
% + 介绍另一贡献
%   + 除了BEV特征这个角度，我们还试图从decoder的query这种细粒度的层面上来促进时序的融合。(decoder是detr这种基于query的)
%   + 我们重用了历史帧的decoder输出的query，将其与当前帧的query相融合。希望历史信息能提供一个更好的先验。
% + 实验结果
%   + 在经典的框架BEVDet4D, BEVDepth4D和BEVFormer上都做了实验，验证了有效性。
%   + 在val上取得了xxx提升
%   + 在test上取得了sota
\begin{figure}[tp]
    \centering
    \includegraphics[width=0.49\textwidth]{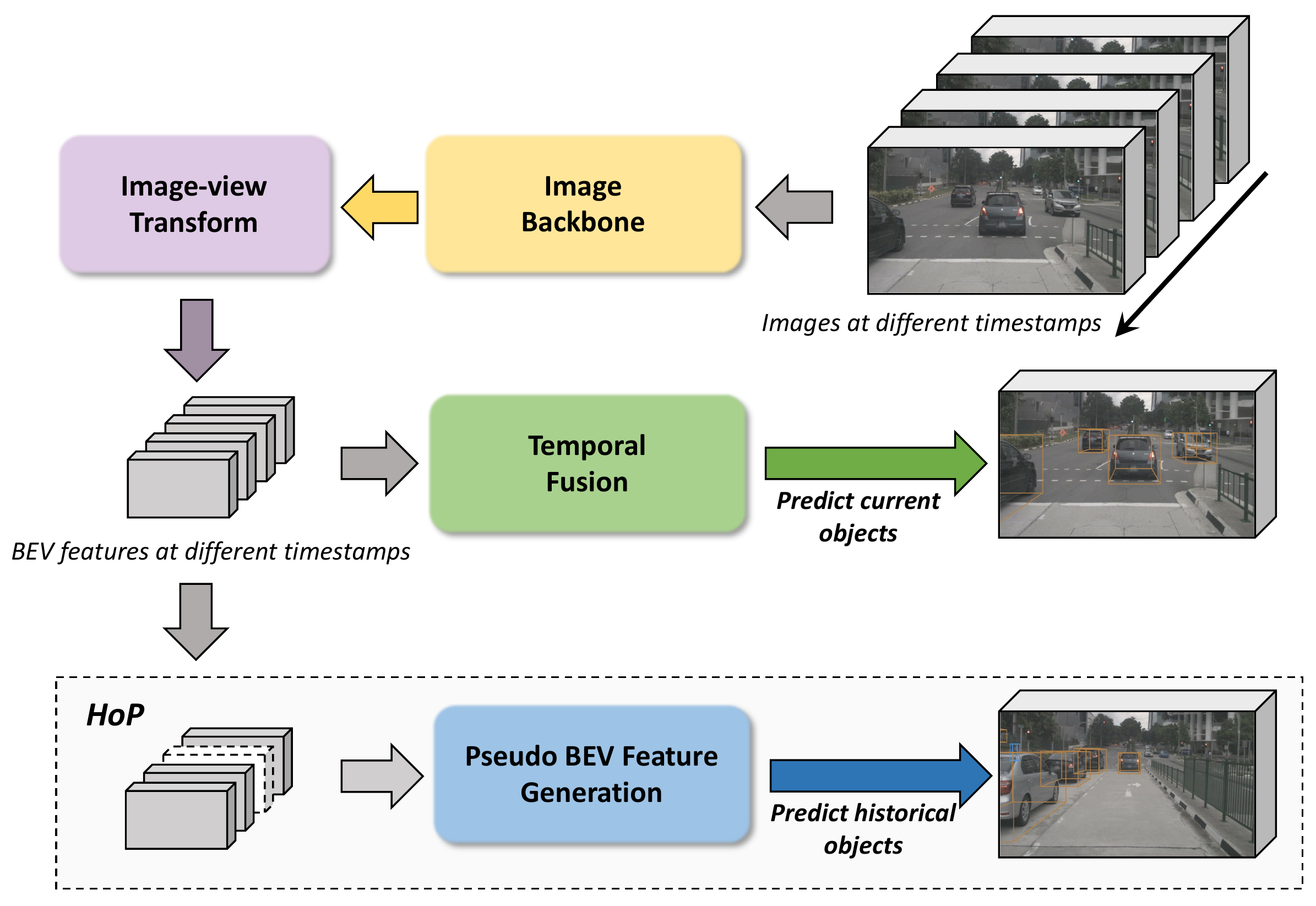}
    \caption{\textbf{Incorporating HoP into 3D object detector with other temporal fusion methods~\cite{bevdet,bevdet4d,bevformerv2,bevstereo,solofusion}.} Our HoP is plug-and-play and complementary to them.}
    \label{fig:framework}
    \vspace{-0.2cm}
\end{figure}

Camera-only 3D detection from multi-view images is a challenging task for autonomous driving and has received increasing attention recently. 
Introducing the Bird's-Eye View (BEV) representation~\cite{lss,bevdepth,bevdet,detr3d,petr,graph_detr3d} with the temporal features aggregation has become the superior design manner for camera-only 3D perception with multi-camera images.
Based on this, a series of detectors~\cite{bevdet4d,bevformer,bevstereo,solofusion,sts,petrv2,sparse4d} delve into the elaborated temporal information fusion methods and achieve significant breakthroughs.
They consider the regions in 3D space and aggregate the image features corresponding to these hypothesis locations from multiple timestamps.
These outstanding advances reveal the great potential of temporal information in camera-only 3D detection.

Beyond these temporal fusion mechanisms, in this paper, we propose a new paradigm for leveraging temporal information to enhance the temporal modeling ability of the 3D object detector.
It's a temporal-based auxiliary task only adopted during training stage, namely Historical Object Prediction (HoP).
Our approach is motivated by the observation that enforcing the detector to capture both the spatial location and temporal motion of objects occurring at historical timestamps can lead to more accurate BEV feature learning.
Specifically, given the current timestamp $t$, we generate a pseudo BEV feature of timestamp $t$-$k$ from its adjacent frames and utilize this feature to predict the object set at $t$-$k$. 
First, we elaborately design short-term and long-term temporal decoders, which can generate the pseudo BEV feature for timestamp $t$-$k$ without the involvement of its corresponding camera images. 
Thanks to the marginal temporal difference between two adjacent frames, the short-term temporal decoder process two adjacent frames of timestamp $t$-$k$ to provide spatial semantics of objects.
The long-term decoder captures long-term motion of the whole frame sequence and contributes to better object motion estimation, which is complementary to spatial localization of the short-term branch.
Subsequently, an additional object decoder is flexibly attached to predict the object targets using the generated pseudo BEV feature. 
% The short-term and long-term temporal decoders are elaborately designed to generate the 
% pseudo BEV feature $\hat{B}_{t-k}$ for timestamp $t$-$k$.
% Thanks to the marginal temporal difference between two adjacent frames, the short-term temporal decoder process adjacent frames of frame $t$-$k$ to provide reliable spatial localization of objects.
% And the long-term decoder captures long-term motion of the whole frame sequence, contributing to better object motion estimation.
% % For the long-term temporal decoder, 
% Note that, the camera images at timestamp $t$-$k$ are not utilized for the long-term and short-term temporal decoder.
% An extra object decoder is flexibly attached to predict the object set $\hat{G}_{t-k}$
%  upon the pseudo $t$-$k$ BEV feature.
% HoP encourages the backbone to learn xx knowledge 
% 介绍HoP的优势xx 1、只在训练使用，不增加测试cost。2、即插即用适合所有BEV的检测器。3、xx
% 讲清楚你和之前temporal的方法是互补的，可以在他们之上用，是辅助训练的。Fig1就是show在他们之上加你的方法的提升。
Note that we only perform HoP during training, thus the proposed method does not introduce extra overheads during inference.
% As a plug-and-play approach, HoP can be easily incorporated into state-of-the-art BEV detection frameworks, including BEVFormer and BEVDet series. 
% Furthermore, the auxiliary HoP approach is complementary to prevalent temporal modeling methods, leading to significant performance gains.
% Extensive experiments are conducted to evaluate the effectiveness of the proposed HoP on the nuScenes dataset.
% We choose the representative methods, including BEVFormer and BEVDet4D-Depth to evaluate our method.
% We only perform HoP during training, thus the proposed method does not introduce extra overheads during inference.
As a plug-and-play approach, HoP can be flexibly incorporated into the state-of-the-art BEV detection frameworks, including BEVFormer~\cite{bevformer} and BEVDet series~\cite{bevdet,bevdet4d,bevdepth}.
Furthermore, the auxiliary HoP approach is complementary to prevalent temporal modeling methods, leading to significant gains.

Extensive experiments are conducted to evaluate the effectiveness of the proposed HoP on the nuScenes dataset~\cite{nuscenes}. 
% For convenience, we utilize BEVFormer and BEVDet4D-Depth to evaluate our method.
We choose the representative methods, including BEVFormer and BEVDet4D-Depth~\cite{bevdet4d,bevdepth} to evaluate our method.
To be specific, we obtain 55.8\% NDS with ResNet-101-DCN and 60.3\% NDS with VoVNet-99  when evaluating HoP on nuScenes \textit{val}.
% As a plug-and-play approach, our methods can be flexibly incorporated into the state-of-the-art BEV detection frameworks, including BEVFormer~\cite{bevformer} and BEVDet series~\cite{bevdet,bevdet4d}, yielding significant performance gains in the 3D object detection task.
Surprisingly, HoP achieves 68.5\% NDS and 62.4\% mAP with ViT-L~\cite{vit} on nuScenes \textit{test}, surpassing all the 3D object detectors on the leaderboard by a large margin.

\begin{figure*}[tp]
    \centering
    \includegraphics[width=0.9\textwidth]{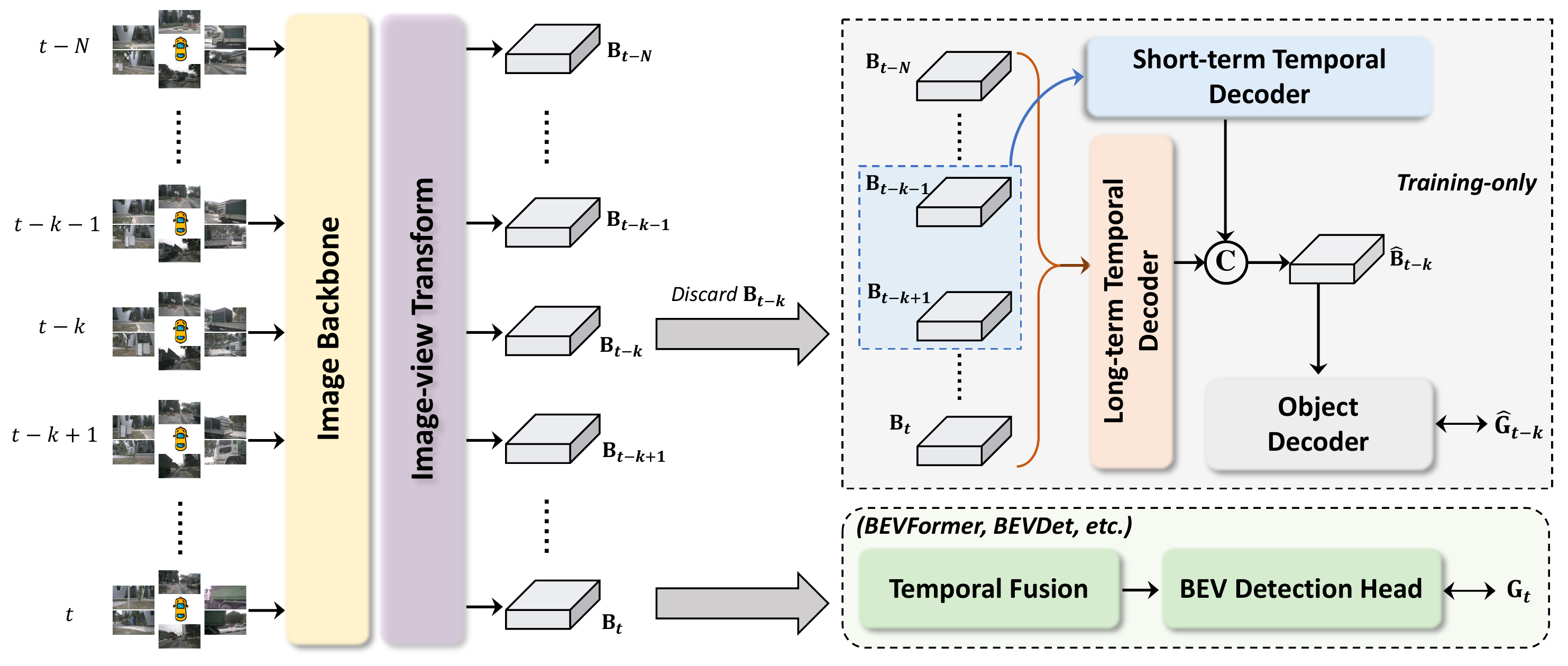}
    \caption{\textbf{Framework of our Historical Object Prediction (HoP).} The auxiliary branches are discarded during evaluation. The symbol \textbf{\copyright} denotes feature concatenation and fusion.}
    \label{fig:framework}
    \vspace{-0.6cm}
\end{figure*}

In conclusion, our contributions can be summarized as follows:
\begin{itemize}
\item
We propose a novel temporal enhanced training scheme, namely Historical Object Prediction (HoP), to encourage more accurate BEV feature learning.
HoP can force the model to capture the spatial semantics and temporal motion of objects in the historical frame during training.
\item
We design a temporal decoder that consists of a short-term decoder and a long-term decoder to provide reliable spatial localization and accurate motion estimation of objects.
\item
We equip the competitive 3D object detector baselines with our approach and yield significant improvements on the nuScenes dataset. 
Surprisingly, HoP with ViT-L achieves 68.5\% NDS and 62.4\% mAP on nuScenes \textit{test}, establishing the new state-of-the-art performance.
\end{itemize}

%% file: 2_related.tex
\section{Related Works}
\label{sec:related}

\subsection{Multi-view 3D Object Detection}
% query-based and lls-based
Modern multi-view methods for 3D object detection could be mainly categorized into two branches, LSS-based~\cite{lss} and query-based methods. 
We will introduce how these two methods differ in feature aggregation and leave the temporal modeling part in the next section.

BEVDet~\cite{bevdet} is a representative method in LSS-based methods. Following the Lift-Splat-Shoot paradigm, the method first explicitly estimates depth for every image pixel, then lifts the 2D features to 3D voxels according to the depth, and finally splats 3D features to BEV features and conducts object detection on it. BEVDepth~\cite{bevdepth} further improves the view transformation module with explicit depth supervision.

Query-based methods typically employ learnable queries to aggregate 2D image features by attention~\cite{transformer, deformable} mechanism. DETR3D~\cite{detr3d} utilizes object queries for predicting 3D positions and projects them back to 2D coordinates to obtain the corresponding features. 
Graph-DETR3D~\cite{graph_detr3d} and Sparse4D~\cite{sparse4d} enhance it respectively with a learnable 3D graph and sparse 4D sampling. 
PETR~\cite{petr} encodes 3D position into image features and therefore directly query with global 2D features. 
BEVFormer~\cite{bevformer} leverages grid-shaped BEV queries to interact with 2D features by deformable attention; on its basis, BEVFormerv2~\cite{bevformerv2} introduces perspective supervision. 
Besides, PolarFormer and Ego3RT~\cite{ego3rt, polarformer} advocates modeling BEV queries in polar coordinates to fit the real-world scenario.

\subsection{Temporal Modeling for Multi-view 3D Object Detection}
The motion information in autonomous driving scenes has been increasingly explored to utilize the temporal cues for improving detection performance in recent 3D perception frameworks.
BEVFormer~\cite{bevformer} proposes the temporal self-attention mechanism to dynamically fuse the previous BEV features by deformable attention~\cite{deformable} in an RNN manner.
BEVDet4D~\cite{bevdet4d} introduces the temporal modeling to lift BEVDet~\cite{bevdet} to spatial-temporal 4D space. 
The 3D position embedding (3D PE) in PETR~\cite{petr} is extended by PETRv2~\cite{petrv2} with the temporal alignment.
BEVStereo~\cite{bevstereo} and STS~\cite{sts} both leverage temporal views for constructing multi-view stereo by an effective temporal stereo method.
SOLOFusion~\cite{solofusion} fully exploits the synergy of short-term and long-term temporal information that is highly complementary as well as achieving state-of-the-art performance.
BEVFormerV2~\cite{bevformerv2} uses the bidirectional temporal encoder to utilize both history and future BEV features, improving the performance by a large margin.
In contrast with aforementioned works, we provide a novel temporal enhanced training paradigm to improve the temporal modeling without additional inference costs.

%% file: 3_method.tex
\section{Method}
\subsection{Overall Architecture}

As illustrated in Figure~\ref{fig:framework}, HoP can be easily incorporated into many multi-view 3D object detectors, \eg, BEVFormer and BEVDet4D-Depth.
We transform the extrinsic parameters of previous frames to the coordinate system of the current ego to perform temporal alignment following BEVDet.

% is built upon the 
% optimized multi-view 3D object detector BEVFormer as illustrated in Figure~\ref{fig:framework}.
% The image features of $6$ cameras are extracted and then lifted into the BEV feature by the BEV spatial encoder.
% Multiple BEV features of different frames are fused by the BEV temporal encoder following~\cite{} but we do not warp these features.
% Instead, we transform the extrinsic parameters of previous frames to the coordinate system of the current ego to perform the temporal alignment.
% To further facilitate the temporal modeling of the BEV detector, we introduce an auxiliary training task to detect the 3D objects of the truncated previous frame.
% Note that the proposed training scheme only introduces extra overheads during training.
% Besides, we revamp the transformer decoder as a temporal multi-stage decoder for better incorporating long-term temporal information from historical queries.

\subsection{Historical Object Prediction}

\noindent\textbf{Pipeline.}
The architecture of our Historical Object Prediction (HoP) is composed of a temporal decoder $\mathcal{T}$ and an object decoder $\mathcal{\hat{D}}$.
Given the BEV feature sequence $\{\mathbf{B}_{t-N}, \cdots, \mathbf{B}_t\}$ that consists of $N$ historical BEV features and the current BEV feature, we denote the  corresponding 3D ground truths as $\{\mathbf{G}_{t-N}, \cdots, \mathbf{G}_t\}$.
If the BEV feature at timestamp $t$-$k$ is discarded, we use the remaining BEV feature sequence $\{\mathbf{B}_{t-N}, \cdots, \mathbf{B}_t\}$ - $\{\mathbf{B}_{t-k}\}$ (denoted as $\mathbf{B}^{rem}$) to predict the 3D objects at timestamp $t$-$k$.
To be specific, we adopt $\mathcal{T}$ and $\mathcal{\hat{D}}$ to transform the remaining BEV features to obtain the 3D predictions $\mathbf{\hat{P}}_{t-k}$:
\begin{equation}
    \label{eq:pred}
    \mathbf{\hat{P}}_{t-k} = \mathcal{\hat{D}}(\mathcal{T}(\{\mathbf{B}_{t-N}, \cdots, \mathbf{B}_t\} - \{\mathbf{B}_{t-k}\})).
\end{equation}
The temporal decoder $\mathcal{T}$ that models both short-term and long-term information is developed to reconstruct the BEV feature at timestamp $t$-$k$.
The object decoder $\mathcal{\hat{D}}$ is attached to generate the predictions $\mathbf{\hat{P}}_{t-k}$ using the pseudo BEV feature.
Considering the ego-motion and temporal alignment, we transform 3D coordinates of $\mathbf{G}_{t-k}$ into the coordinate system of current frame $t$ (denoted as $\mathbf{\hat{G}}_{t-k}$).
The final training objective is to minimize the differences between $\mathbf{\hat{P}}_{t-k}$ and $\mathbf{\hat{G}}_{t-k}$.
% \noindent\textbf{truncated temporal encoder.}
% Given the mask index $k$ and truncated BEV feature $\mathbf{B}_{t-k}$, 

The input to the temporal decoder is the remaining set of BEV features consisting of (i) short-term BEV features $\{\mathbf{B}_{t-k-1}, \mathbf{B}_{t-k+1}\}$, and
(ii) long-term BEV features $\{\mathbf{B}_{t-N}, \cdots, \mathbf{B}_t\}$ - $\{\mathbf{B}_{t-k}\}$. 
We first add temporal positional embeddings to BEV features in this full set.
% to introduce the temporal location information. 
Our temporal decoder reconstructs the feature $\mathbf{\hat{B}}_{t-k}$ by utilizing the spatial semantics and temporal cues of the BEV feature sequence.
More specifically, we present a temporal decoder that consists of two separate branches with their special expertise to capture short-term and long-term information.
The learned strong spatial-temporal representation can help the network better estimate the object location in frame at timestamp $t$-$k$.

\noindent\textbf{Short-term temporal decoder.}
% the temporal difference between the reference and
% source frame varies significantly over different pixels, depths, cameras, and ego-motion.
% The marginal temporal difference between the adjacent frames varies marginally thanks to the time interval.
Thanks to the high temporal correlation between adjacent frames, the short-term temporal decoder only operates on the adjacent BEV feature set $\{\mathbf{B}_{t-k-1}, \mathbf{B}_{t-k+1}\}$ (denoted as $\mathbf{B}^{adj}$), building a detailed spatial representation in BEV space.
We first define a grid-shaped learnable short-term BEV queries $\mathbf{Q}^{short}_{t-k} \in \mathbb{R}^{H \times W \times C}$, where $H$ and $W$ refer to the spatial shape of the BEV plane.
The short-term BEV queries aggregate the spatial information and model the short-term motion from $\mathbf{B}_{t-k-1}$ and $\mathbf{B}_{t-k+1}$ through the short-term temporal attention, which can be formulated as:
\begin{equation}
    \label{eq:short_term}
    \mathbf{\hat{B}}^{short}_{t-k} = \sum_{\mathbf{V} \in \mathbf{B}^{adj}} \mathrm{DeformAttn}(\mathbf{Q}^{short}_{t-k, p},p,\mathbf{V}),
\end{equation}
where $p$ is the spatial index in the BEV plane, $\mathrm{DeformAttn}(q,p,x)$ refers to the deformable attention~\cite{deformable} with query $q$, reference point $p$ and input features $x$.
We further feed $\mathbf{\hat{B}}^{short}_{t-k}$ into a feed-forward network~\cite{transformer} and obtain the output of this short-term branch.
However, it is still challenging to precisely construct the temporal relations of the same objects between $\mathbf{B}_{t-k}$ and other BEV features with only two adjacent frames.

\noindent\textbf{Long-term temporal decoder.}
The long-term temporal processes the whole remaining BEV set to perceive the motion clues over long-term history, which increases the localization potential~\cite{solofusion} and contributes to more accurate localization.
Therefore, these two branches in the temporal decoder are complementary to each other.
As for long-term motion, it is intuitive that we only focus on the \textit{spatial motion} of the same objects in the bird’s eye view, ignoring the height information of objects. 
For most 3D object detectors with 2D BEV features, \eg, BEVFormer and BEVDet, the height information has been flattened to the feature embeddings.
% {\color{blue} delete other methods. given xxx, It is intuitive that we ignore the height information and only focus xxx in the long-term branch}
Accordingly, we first employ a channel reduction operation to the input set $\mathbf{B}^{rem}$ to prune the height information and  achieve better training efficiency. 
Given the learnable long-term BEV queries $\mathbf{Q}^{long}_{t-k} \in \mathbb{R}^{H \times W \times C/r}$ and reduction layer with parameter $\mathbf{W}^{r} \in \mathbb{R}^{C \times C/r}$, we can capture the long-term dependencies as:
\begin{equation}
    \label{eq:long_term}
    \mathbf{\hat{B}}^{long}_{t-k} = \sum_{\mathbf{V} \in \mathbf{B}^{rem}} \mathrm{DeformAttn}(\mathbf{Q}^{long}_{t-k, p},p,\mathbf{V}\mathbf{W}^{r}),
\end{equation}
where $r$ denotes the reduction ratio and is set as 4 by default.
After the feed-forward network, the long-term decoder output the BEV features.
Finally, we concatenate the short-term and long-term BEV features and perform feature fusion by 3$\times$3 convolution.
% Note that we only focus on the spatial differences between frames in the bird’s eye view
% Note that we first employ a squeeze operation to
% capturing spatial motion without building a
% detailed spatial representation

\noindent\textbf{Object decoder.}
The reconstructed BEV feature $\mathbf{\hat{B}}_{t-k}$ is further processed by a lightweight object decoder.
% Note that the decoder is used to generate 3D predictions upon the BEV feature and thus can be flexibly implemented by any current BEV detection head.
% For example, we choose CenterPoint as the object decoder due to its simplicity.
% The implementation of this object decoder is flexible. 
The decoder is used to generate 3D predictions $\mathbf{\hat{P}}_{t-k}$ upon the BEV feature and thus its implementation is flexible.
Variants of BEV detection heads are an obvious choice and we will consider multiple instantiations in our experiments.
After obtaining $\mathbf{\hat{P}}_{t-k}$, we should align the coordinates between learning targets $\mathbf{G}_{t-k}$ and the final predictions $\mathbf{\hat{P}}_{t-k}$.
For clarity, we first denote the ego coordinate as $e(t)$ at frame $t$.
In our implementation, we transform the extrinsic parameters of previous frames to the coordinate system of the current ego, thus BEV features and predictions at different timestamps share the same $e(t)$.
To simplify the learning targets, 3D coordinates of $\mathbf{G}_{t-k}$ should be transformed into $e(t)$.
Considering the ego motion, we convert $\mathbf{G}_{t-k}$ to $\mathbf{\hat{G}}_{t-k}$ by ego transformation matrix as:
\begin{equation}
    \label{eq:transform}
    \mathbf{\hat{G}}_{t-k} = \mathbf{T}^{e(t)}_{e(t-k)}\mathbf{G}_{t-k},
\end{equation}
where $\mathbf{T}^{e(t)}_{e(t-k)}$ is the transformation matrix from the source coordinate system $e(t-k)$ to the target coordinate system $e(t)$.
% Therefore, the coordinate systems of learning targets $\mathbf{G}_{t-k}$ (ego at $t$) and final predictions $\mathbf{\hat{P}}_{t-k}$ (ego at $t$-$k$) are inconsistent.
% Considering the ego motion

% After obtaining $\mathbf{\hat{P}}_{t-k}$, we should align the coordinates between learning targets $\mathbf{G}_{t-k}$ and the final predictions $\mathbf{\hat{P}}_{t-k}$. 

% Considering the ego motion, we should align the coordinates between learning targets $\mathbf{G}_{t-k}$ and the final predictions $\mathbf{\hat{P}}_{t-k}$.
% 3D coordinates of the ground truth in frame $t$-$k$ should be transformed into the coordinate system of the current frame $t$.
% Accordingly, we convert $\mathbf{G}_{t-k}$ to $\mathbf{\hat{G}}_{t-k}$ by ego transformation matrix and minimize the difference between transformed $\mathbf{\hat{G}}_{t-k}$ and predictions $\mathbf{\hat{P}}_{t-k}$.

% \noindent\textbf{Training objective.}
% Considering the ego motion, 3D coordinates of the ground truth in frame $t$-$k$ should be transformed into the coordinate system of the current frame $t$.
% Accordingly, we convert $\mathbf{G}_{t-k}$ to $\mathbf{\hat{G}}_{t-k}$ by ego transformation matrix and the training objective of the temporal truncated objects detection task can be illustrated as:
% \begin{equation}
%     \label{eq:objective}
%     \mathcal{L}_{pred} = \mathcal{L}_{3d}(\mathbf{\hat{P}}_{t-k}, \mathbf{\hat{G}}_{t-k}),
% \end{equation}
% where $\mathcal{L}_{3d}$ is the 3D object detection loss.

\begin{figure}[tp]
\centering 
\includegraphics[width=0.45 \textwidth]{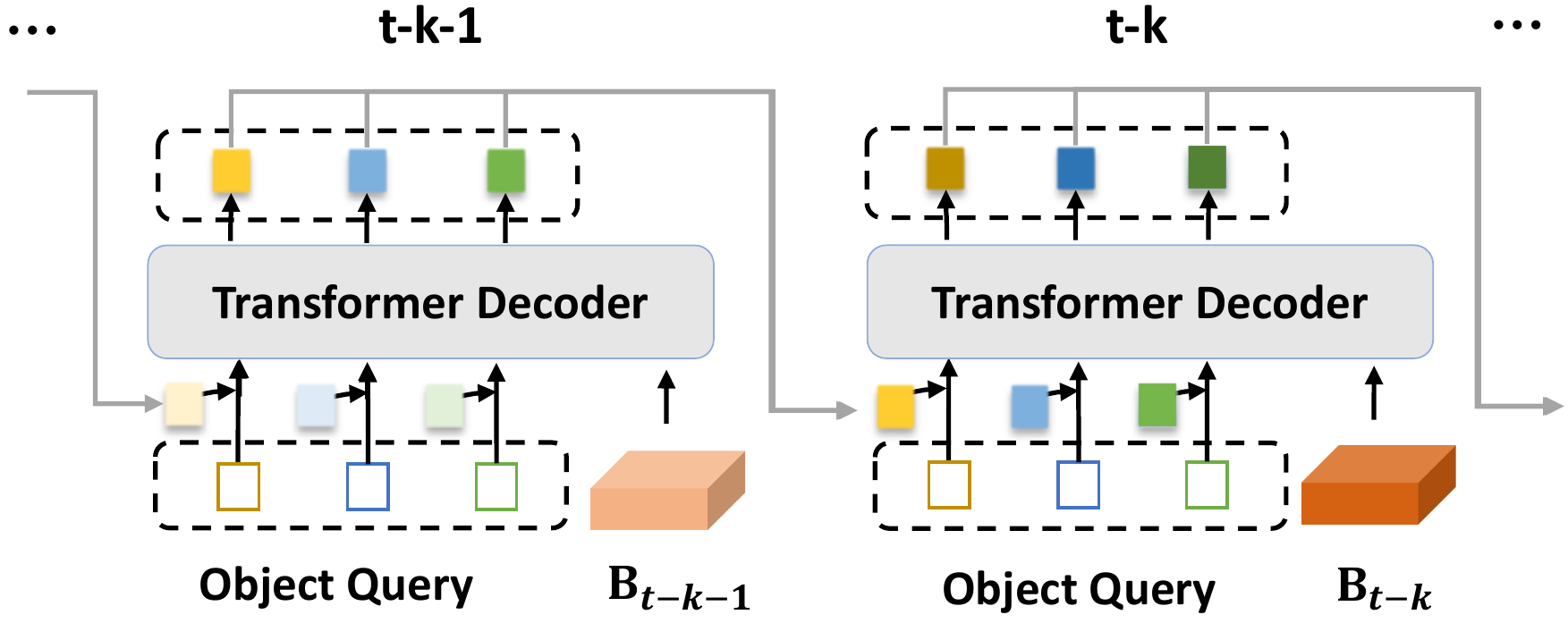}
\caption{Illustration of the temporal multi-stage decoder. Object queries are enhanced by their counterparts derived from the last frame, forming a recurrent connection.} 
\label{Fig.connection-form} 
\end{figure}

\input{main_val}
\input{main_test}
% \subsection{Temporal Multi-stage Decoder}
\subsection{Historical Temporal Query Fusion}
\label{method: query fusion}
Apart from the HoP method that exploits temporal information in BEV feature level, we also explore query-level temporal modeling in query-based methods, \eg, BEVFormer, by fusing historical object queries into current queries.
% For query-based methods, to further explore the use of temporal information, we propose to fuse historical object queries into current queries to leverage its instance-level information of history frames.
This fusion step provides current queries with an initialization of history perception, forming a refining process.

% we first obtain temporal enhanced BEV feature $\mathbf{\bar{B}}_{t-k}$ from $\{\mathbf{B}_{t-N}, \cdots, \mathbf{B}_{t-k}\}$ with BEV temporal encoder {\color{blue} cite?}. Then we perform detection with the pre-defined object queries $\mathbf{O}$ as follows:
The BEV feature sequence $\{\mathbf{B}_{t-N}, \cdots, \mathbf{B}_t\}$ is inferred in the time order. 
For the timestamp $t$-$k$, we perform detection with the pre-defined object queries $\mathbf{O}$ as follows:
\begin{equation}
    \label{eq:qpred}
    \mathbf{\bar{O}}_{t-k} = \mathcal{D}(\mathbf{B}_{t-k},\mathbf{O})
\end{equation}
where $\mathcal{D}$ is the BEV decoder and $\mathbf{\bar{O}}_{t-k}$ is its output.

Each object query learns to specialize in certain areas in the BEV plane, so the output queries contain rich semantic information from the regions. When detecting on the current frame, we refine these output historical queries instead of starting from the pre-defined ones to make the learning process easier. For example, if an object query detects a still object in the last frame, its historical counterpart could help it quickly locate the object without searching on its area in the BEV plane again. On the other hand, historical information from a moving object also contributes to localizing and detecting its velocity.

As shown in Figure \ref{Fig.connection-form}, to infer on the BEV feature $\mathbf{B}_{t-k}$, we first collect historical object queries $\mathbf{O}^{his}_{t-k}$ as follows:
\begin{equation}
    \label{eq:hist-queries}
    \mathbf{O}^{his}_{t-k} = \mathrm{MLP}(\mathbf{\bar{O}}_{t-k-1})
\end{equation}
where $\mathrm{MLP}$ is a multi-layer perceptron with two linear layers and an expansion ratio of 4, which is used to fit the history query into the current timestamp. Note that we only use historical queries from the last timestamp because we hypothesize that queries from more distant history possess little information overlap with currently visible objects.

Then we merge them with the pre-defined object queries $\mathbf{O}$ as Equation \ref{eq:sum},
\begin{equation}
    \label{eq:sum}
    \mathbf{\bar{O}}_{t-k} = \mathcal{D}(\mathbf{B}_{t-k},\mathbf{O}+\mathbf{O}^{his}_{t-k})
\end{equation}
Considering the first shown-up objects in the current frame, the merge process takes both historical queries and pre-defined queries into account.
% e.g., a cross-attention layer or a simple addition operation.

% \begin{table}[t]
%     \centering\setlength{\tabcolsep}{6pt}
%     \footnotesize
%     \renewcommand{\arraystretch}{1.2}
%     \resizebox{1\linewidth}{!}
%     {
%         \begin{tabular}{l|c|c|c}
%         \shline
%         Method &  Temporal Agg. & Task &  Infer. Cost \\
%         \hline
%         \multirow{1}*{BEVDet4D} & align \& concat& temporal info. agg.&\cmark \\
%         \multirow{1}*{BEVFormer} & attention & temporal info. agg.&\cmark \\
%         \multirow{1}*{BEVStereo} & stereo matching& depth estimation&\cmark \\
%         \hline
%         \multirow{2}*{HoP} & \multirow{2}*{attention} & historical object &\multirow{2}*{\xmark} \\
%          &  & prediction& \\
%         \shline
%         \end{tabular}
%     }
%     \vspace{-2mm}
%     \caption{\small{Comparison with other methods in the exploitation of temporal information.}
%     }
%     \label{tab:comparison}
%     \vspace{-5mm}
% \end{table}

% \subsection{Comparison with Other Methods}
% Table \ref{tab:comparison} compares HoP with other methods in the exploitation of temporal information. 
% Our method leverages historical object prediction as an auxiliary task to enhance the temporal modeling ability of 3D detector. 
% Hence, it introduces no inference cost and could serve as a plug-and-play approach.

%% file: main_val.tex
\begin{table*}[tp]
    \centering
    % \small
    \addtolength{\tabcolsep}{-3.5pt}
    \renewcommand{\arraystretch}{1.2}
    \begin{tabular}{l|c|c|c|c|cc|ccccc}
    \Xhline{1.0pt}
    Method & Backbone & Query-based & LiDAR & Epoch & NDS$\uparrow$ & mAP$\uparrow$ & mATE$\downarrow$ & mASE$\downarrow$ & mAOE$\downarrow$ & mAVE$\downarrow$ & mAAE$\downarrow$ \\
    % \hline
    % BEVDet4D & ResNet50 &  &  & 90$^\dagger$ & 0.457 & 0.332 & 0.703 & 0.278 & 0.495 & 0.354 & 0.206 \\
    % BEVDepth & ResNet50 &  & \cmark & 90$^\dagger$ & 0.475 & 0.351 & 0.639 & 0.267 & 0.479 & 0.428 & 0.198 \\ 
    % BEVStereo & ResNet50 &  & \cmark & 90$^\dagger$ & 0.500 & 0.372 & 0.598 & 0.270 & 0.438 & 0.367 & 0.190 \\ 
    % SOLOFusion & ResNet50 &  & \cmark & 90$^\dagger$ & 0.534 & 0.427 & 0.567 & 0.274 & 0.411 & 0.252 & 0.188 \\
    % PETRv2 & ResNet50 & \cmark &  & 24 & 0.494 & 0.398 & 0.690 & 0.273 & 0.467 & 0.424 & 0.195 \\
    % BEVFormerV2 & ResNet50 & \cmark &  & 24 & 0.498 & 0.388 & 0.679 & 0.276 & 0.417 & 0.403 & 0.189 \\
    % \rowcolor[gray]{.9}
    % our & ResNet50 & \cmark &  & 24 &{\color{blue} consistent} & & & & & & \\
    \hline
    BEVDepth~\cite{bevdepth} & R101-DCN$^\ddagger$ &  & \cmark & 90$^\dagger$ & 0.538 & 0.418 & - & - & - & - & - \\
    UVTR~\cite{uvtr} & R101-DCN$^\ddagger$ & \cmark &  & 24 & 0.483 & 0.379 & 0.731 & 0.267 & 0.350 & 0.510 & 0.200 \\
    BEVFormer~\cite{bevformer} & R101-DCN$^\ddagger$ & \cmark &  & 24 & 0.517 & 0.416 & 0.673 & 0.274 & 0.372 & 0.394 & 0.198 \\
    PETRv2~\cite{petrv2} & R101-DCN$^\ddagger$ & \cmark &  & 24 & 0.524 & 0.421 & 0.681 & 0.267 & 0.357 & 0.377 & 0.186 \\
    PolarFormer~\cite{polarformer} & R101-DCN$^\ddagger$ & \cmark &  & 24 & 0.528 & 0.432 & 0.648 & 0.270 & 0.348 & 0.409 & 0.201 \\
    Sparse4D~\cite{sparse4d} & R101-DCN$^\ddagger$ & \cmark &  & 24 & 0.541 & 0.436 & 0.633 & 0.279 & 0.363 & \textbf{0.317} & \textbf{0.177} \\
    \rowcolor[gray]{.9}
    HoP-BEVFormer & R101-DCN$^\ddagger$ & \cmark &  & 24 & \textbf{0.558} & \textbf{0.454} & \textbf{0.565} & \textbf{0.265} & \textbf{0.327}  & 0.337 & 0.194 \\
    \Xhline{1.0pt}
    \end{tabular}
    \vspace{-1mm}
    \caption{Comparison of recent works on the nuScenes val set. $\dagger$ indicates methods with CBGS which will elongate 1 epoch into 4.5 epochs. $\ddagger$ notes the backbone is initialized from FCOS3D~\cite{fcos3d} backbone. LiDAR: trained with LiDAR supervision.}
    \vspace{-2mm}
    \label{tab:main_val}
    \normalsize
\end{table*}

%% file: main_test.tex
\begin{table*}[t]
    \centering
    % \small
    \addtolength{\tabcolsep}{-3.0pt}
    \renewcommand{\arraystretch}{1.2}
    \begin{tabular}{l|c|c|c|c|cc|ccccc}
    \Xhline{1.0pt}
    Method & Backbone & Query-based & LiDAR & Epoch & NDS$\uparrow$ & mAP$\uparrow$ & mATE$\downarrow$ & mASE$\downarrow$ & mAOE$\downarrow$ & mAVE$\downarrow$ & mAAE$\downarrow$ \\
    \hline
    BEVDepth$^*$~\cite{bevdepth} & V2-99 &  & \cmark & 90$^\dagger$ & 0.600 & 0.503 & \textbf{0.445} & 0.245 & 0.378 & 0.320 & 0.126 \\
    % BEVStereo~\cite{bevstereo} & V2-99 &  & \cmark & 90$^\dagger$ & 0.610 & 0.525 & 0.431 & 0.246 & 0.358 & 0.357 & 0.138 \\
    DETR3D~\cite{detr3d} & V2-99 & \cmark &  & 24 & 0.479 & 0.412 & 0.641 & 0.255 & 0.394 & 0.845 & 0.133 \\
    UVTR~\cite{uvtr} & V2-99 & \cmark &  & 24 & 0.551 & 0.472 & 0.577 & 0.253 & 0.391 & 0.508 & 0.123 \\
    BEVFormer~\cite{bevformer} & V2-99 & \cmark &  & 24 & 0.569 & 0.481 & 0.582 & 0.256 & 0.375 & 0.378 & 0.126 \\
    PETRv2~\cite{petrv2} & V2-99 & \cmark &  & 24 & 0.582 & 0.490 & 0.561 & \textbf{0.243} & 0.361 & 0.343 & 0.120 \\
    Sparse4D~\cite{sparse4d} & V2-99 & \cmark &  & 24 & 0.595 & 0.511 & 0.533 & 0.263 & 0.369 & \textbf{0.317} & 0.124 \\  
    % UVTR V2-99 900 × 1600 ✓ ✗ 0.472 0.551 0.577 0.253 0.391 0.508 0.123
    % BEVFormer V2-99 900 × 1600 ✓ ✗ 0.481 0.569 0.582 0.256 0.375 0.378 0.126
    % BEVDet4D Swin-B 900 × 1600 ✗ ✓ 0.451 0.569 0.511 0.241 0.386 0.301 0.121
    % PolarFormer V2-99 900 × 1600 ✓ ✗ 0.493 0.572 0.556 0.256 0.364 0.439 0.127
    % PETRv2 GLOM-like 640 × 1600 ✗ ✗ 0.512 0.592 0.547 0.242 0.360 0.367 0.126
    % BEVDepth ConvNeXt-B 640 × 1600 ✗ ✗ 0.520 0.609 0.445 0.243 0.352 0.347 0.127
    % BEVStereo V2-99 640 × 1600 ✓ ✗ 0.525 0.610 0.431 0.246 0.358 0.357 0.138
    \rowcolor[gray]{.9}
    HoP-BEVFormer & V2-99 & \cmark &  & 24 & \textbf{0.603} & \textbf{0.517} & 0.501 & 0.245 & \textbf{0.346} & 0.362 & \textbf{0.105} \\
    \Xhline{1.0pt}
    \end{tabular}
    \vspace{-1mm}
    \caption{Comparison of recent works on the nuScenes test set. $*$ indicates test-time augmentation. VoVNet-99 (V2-99)~\cite{vov} was pre-trained on the depth estimation task with extra data~\cite{dd3d}.}
    \label{tab:main_test}
    \normalsize
    \vspace{-3mm}
\end{table*}

%% file: 4_results.tex
\section{Experiments}
\label{sec:results}

\input{sota}
\input{component}
\input{new_baseline}

\subsection{Dataset and Metrics}
We validate our method on the popular nuScenes\cite{nuscenes} dataset. The nuScenes dataset contains 1000 driving scenes in total, which are split into 700, 150, and 150 respectively for training, validation, and testing. Each scene lasts for 20 seconds and annotations of 3D boxes are supplied every 0.5s in the keyframe. There are $6$ cameras surrounding the vehicle, providing images covering the whole 360-degree FOV. As for evaluation, we adopt the standard metrics for nuScenes detection task. The final nuScenes detection score (NDS) is a weighted sum of mean Average Precision (mAP) and five True Positive metrics including 
% mAP is averaged on 4 matching thresholds and excludes the area where recalls or precisions are lower than 0.1. 
Average Translation Error (ATE), Average Scale Error (ASE), Average Orientation Error (AOE), Average Velocity Error (AVE), and Average Attribute Error (AAE).

\subsection{Implementation Details}
We mainly implement our method on the optimized version of BEVFormer~\cite{bevformer} (denoted as BEVFormer-opt), where we generate the current BEV feature with resolution 200$\times$200 by 6 BEV spatial encoder layers and fuse the current and 4 previous BEV features by a BEV temporal encoder.
% The fused features are further fed into the BEV temporal encoder that consists of two 2D residual blocks.
All models are trained with a mini-batch of 8, an initial learning rate of 0.0002, and a weight decay of 0.01 for 24 epochs without CBGS~\cite{cbgs}.
% The learning rate is decreased by a factor of $10$ at the 20-th epoch.
Unless otherwise specified, we use ResNet-50~\cite{resnet} pre-trained on COCO~\cite{coco} as the image backbone, and the image size is processed to 1408$\times$512.
Both image and BEV augmentation are adopted following BEVDet~\cite{bevdet}.
Besides, we also adopt the BEVDet4D-Depth~\cite{bevdet4d} as baseline models and follow the original settings.
% We sample $4$ previous frames for temporal modeling and no future frame is used.

\input{ablation}

\begin{figure*}[thbp]

\centering
\includegraphics[width=1.0 \textwidth]{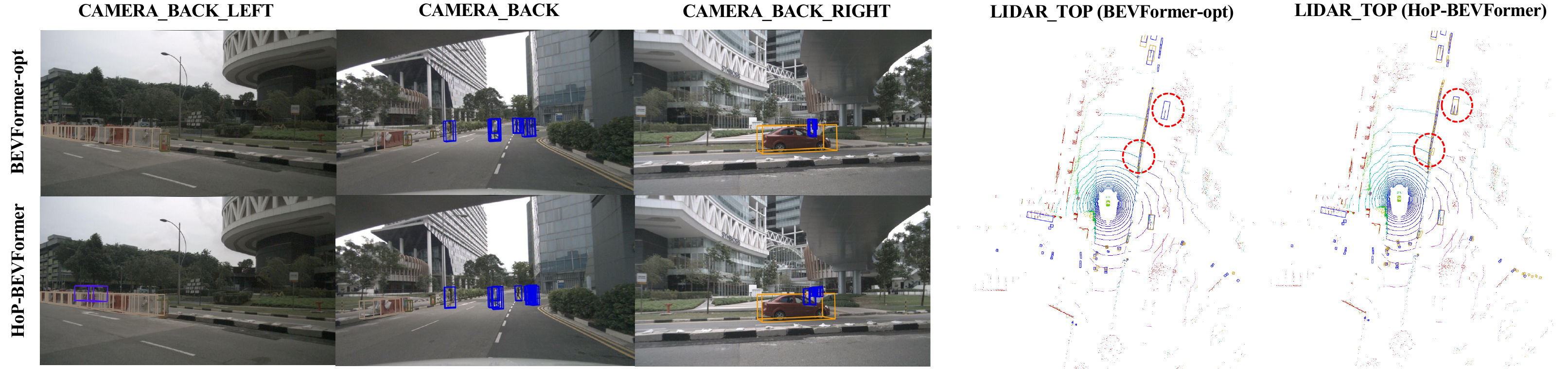}
\caption{Visualization result of BEVFormer-opt and HoP-BEVFormer on nuScenes val set. The predicted boxes are marked in yellow, while the ground truth boxes are marked in blue in the LIDAR\_TOP figure.} 
\label{Fig.vis-baseline} 

\vspace{-6mm}
\end{figure*}

\subsection{Main Results}
To compare with previous state-of-the-art 3D object detection methods, we adopt ResNet101-DCN~\cite{resnet,dcn} initialized from FCOS3D~\cite{fcos3d} and VoVNet-99~\cite{vov,vovv2} initialized from DD3D~\cite{dd3d} checkpoint.
% Besides, we use ViT-L~\cite{vit} which is initialized from FCOS3D to verify the scalability of our training scheme.
We also enlarge the input resolution to 1600$\times$640 and use 1500 object queries.
The historical temporal query fusion is employed in Table~\ref{tab:main_val} and~\ref{tab:main_test}.

We report the results of our HoP with the ResNet101-DCN backbone on nuScenes \textit{val} in Table~\ref{tab:main_val}.
Surprisingly, our method outperforms BEVDepth with CBGS strategy and depth supervision from LIDAR over +2.0\% NDS and +3.6\% mAP under comparable complexity levels.
Besides, HoP achieves the best performance compared with other competitive query-based object detectors, \eg, PETRv2 and PolarFormer.
We also compared HoP with other state-of-the-art algorithms on nuScenes \textit{test}, and still
obtain the best performance of 60.3\% NDS and 51.7\% mAP.
% It is worth noting that HoP achieves slightly better performance than BEVDepth with test-time augmentation, but the improvements in mAOE
% specific, the improvements in mAOE and mAVE are significant: mAOE and mAVE are improved by 3.6% and 4.8%,

To show the scalability of our model to stronger backbones, we apply HoP to BEVDet4D-Depth with ViT-L~\cite{vit} as the backbone.
The ViT-L we employ is initialized with the Co-DETR~\cite{codetr} pre-trained on the 2D detection benchmark Objects365~\cite{objects365}.
We use 8 frames into the future to provide additional information from the future.
To reduce the training time, we first train the ViT-L using FCOS3D on the nuScenes \textit{trainval} for 36 epochs.
Then we only train HoP-BEVDet4D-Depth for 5 epochs with CBGS.
% Without test-time augmentation, HoP achieves 68.2\% NDS and 61.6\% mAP, surpassing the previous state-of-the-art BEVDet-Gamma~\cite{bevdet} with multi-scale testing by a large margin of +1.8\% NDS and +3.0\% mAP.
Without test-time augmentation, HoP pushes the current best result to a new record of 68.5\% NDS and 62.4\% mAP on nuScenes \textit{test}.

\subsection{Ablation Studies}
Unless stated otherwise, we conduct the ablation experiments on BEVFormer-opt with a ResNet-50 backbone.
We follow the original settings when adopting BEVDet4D as the baseline.
% More ablations can be found in the supplementary materials.

\noindent\textbf{Effectiveness of each component.} 
We perform a component-wise ablation to thoroughly analyze the effect of each component in Table \ref{tab:component}. 
Surprisingly, incorporating the historical temporal prediction into the optimized BEVFormer brings significant performance gains (+1.8\% NDS, +1.4\% mAP) over the baseline.
The overall improvements hold when we apply the training strategy to BEVDet4D-Depth, achieving +1.6\% NDS and +1.4\% mAP improvements, respectively.
Alternatively, the temporal information provided by historical queries also contributes to the performance improvements for BEVFormer, where a +1.2\% NDS gain is observed. 
Without detaching the gradients of historical BEV features in the last two layers of BEV encoder, the model can be further improved to 53.9\% NDS.
These results prove the effectiveness and generalization of our HoP.

\noindent\textbf{Optimized BEVFormer baseline.} 
Table \ref{tab:tricks} illustrates the critical designs of our optimized BEVFormer baseline.
We can observe setting the dropout rate within the transformer as 0 significantly improves the performance and look forward twice~\cite{dino} is an essential factor to improve mAP.
Besides, we replace the heavy backbone with the ResNet50 and decrease the input resolution to 1408$\times$512 to improve the training efficiency of BEVFormer.
% At the same time, we also introduce step learning rate decay and data augmentation 
% to achieve better trade-offs between accuracy and complexity.
At the same time, we also introduce step learning rate decay and data augmentation, which includes BEV augmentation, random flipping, cropping, and rotation, to achieve better performance without inference costs increase.
% The temporal modeling in the original temporal self-attention layer is removed and we insert a BEV temporal encoder after the spatial encoder.
% Replacing the original temporal modeling with the BEV temporal encoder and using
% 4 previous frames, is crucial in the later promotion process.
Decoupling the BEV temporal encoder from the spatial encoder and using 4 previous frames, is crucial in the later promotion process.
Combining these aforementioned modifications on BEVFormer results in our optimized baseline BEVFormer-opt.
Finally, we achieve comparable performance with the original BEVFormer-R101-DCN while running much faster and requiring fewer parameters.

\noindent\textbf{Temporal decoder design.} 
Table~\ref{tab:temporal_decoder} illustrates the key designs of our temporal decoder for prediction targets prediction. 
The performance is always improved when using the structure of one individual decoder alone, reaching 52.9\% NDS and 52.7\% NDS with the short-term and the long-term decoder, respectively.
We achieve the best performance with 53.1\% NDS and 42.8\% mAP when both short-term and long-term decoders are adopted, demonstrating the complementarity between these two branches.

\noindent\textbf{Object decoder design.}
The object decoder detects the 3D targets with the reconstructed features, thus it can be flexibly designed, as studied in Table~\ref{tab:obj_decoder}.
We find these detection heads can bring consistent gains.
For example, ATSS~\cite{atss} is extended to 3D object detection and brings 1.6\% NDS and 1.2\% mAP gains.
% our training scheme is not sensitive to the choice of object decoder since all instantiations~\cite{atss} bring consistent gains.
We choose CenterPoint~\cite{centerpoint}, which achieves the best performance as the default object decoder.

\noindent\textbf{Prediction target.} 
% future, 0, 1, 2
% We compare two prediction prediction targets in Table~\ref{tab:pred_target}.
We also study the design that only predicts pseudo BEV features at timestamp $t$-$k$ supervised by $\mathbf{B}_{t-k}$ generated by the BEV encoder in a self-distillation manner.
As presented in Table~\ref{tab:pred_target}, the feature supervision signals only slightly improve the baseline while the 3D objects supervision boosts the performance by a large margin.
In a summary, explicit supervision from 3D objects in the BEV space is crucial to HoP.

\begin{figure*}[thbp]
% \begin{figure}[thbp]
\centering
\includegraphics[width=0.8 \textwidth]{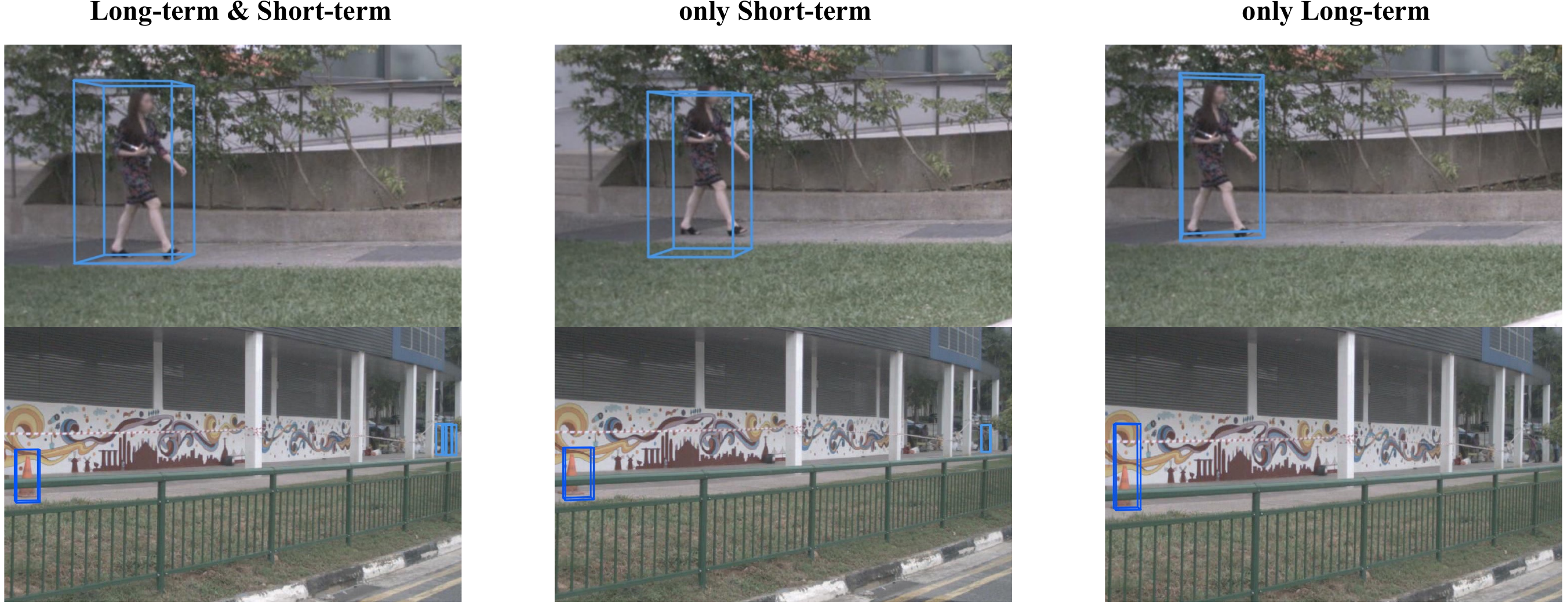}
\caption{Visualization results of object decoder with three variants of the temporal decoder.} 
\label{Fig.vis-ls} 
% \end{figure}
\vspace{-6mm}
\end{figure*}

\noindent\textbf{Choice of prediction index $k$.}
The selection of prediction index $k$ is a critical design of our method and is ablated in Table~\ref{tab:trunc_index}.
We first study whether the model can benefit from predicting the 3D objects in the future frame, \eg, frame $t$+$1$ ($k=-1$). 
Interestingly, this future 3D object prediction also brings non-trivial gains but performs inferior to history prediction.
% This demonstrates the improvements stem from our truncated temporal prediction and
% improvements demonstrate the effectiveness of our truncated temporal prediction 
The relatively weaker gains can be explained by the difference between a history prediction task and a future prediction task: it is much more challenging to predict 3D objects in the future frame at timestamp $t$+$1$ with only history frames from timestamp $t$-$N$ to $t$.
Predicting objects in frame $t$-$k$ with both history frames from timestamp $t$-$N$ to $t$-$k$-1 and future frames from timestamp $t$-$k$+1 to $t$ is much easier.
% A challenging auxiliary task can be less relevant for detecting objects in the current frame, disturbing the learning progress.
A challenging auxiliary task can disturb the learning progress and slightly decrease the performance gains.
When predicting the objects of history frame $t$-$k$, we do not detach the feature maps of frame $t$-$k$+$1$ to allow the gradients back propagate through the both short-term and long-term temporal decoder.
Therefore, we observe the method imposes a significant increase in GPU memory for $k>1$ since the features at timestamp $t$ and $t$-$k$+1 are not detached.
% To achieve the best trade-offs between 
% We observe predicting objects of different history frames achieve similar performance while differing markedly in the training memory consumption.
% Predicting the objects in $t-1$ imposes a significant decrease in GPU memory compared with other cases since we generally unlocks the gradients of 
% In contrast to $k=1$, predicting objects in other historical frames unlocks the gradients of frame $t$ and $t-k+1$ to allow the gradients back propagate through the object decoder.
Accordingly, we choose $k$ as 1 since it achieves the best trade-offs between accuracy and training costs.

\noindent\textbf{Training time.}
We present the training speed (seconds per training iteration) under different settings in Table~\ref{tab:cost}.
This training speed is evaluated with 8 NVIDIA A100 GPUs.
Compared with the baseline, HoP introduces an additional temporal decoder and object decoder during training, leading to an inevitable increase in training time.
Thanks to the lightweight design of our temporal decoder and object decoder, we only observe a 20\% increase (from 1.6 seconds to 1.9 seconds) in training time for BEVFormer baseline.
When the gradients of historical BEV features are not detached in the last two layers of BEV encoder, the training time is increased by 30\%.
As for BEVDet4D-Depth, the BEV feature resolution and dimension is smaller than BEVFormer.
Therefore, the additional training costs brought by HoP are \textit{negligible} (from 4.5 seconds to 4.7 seconds).

\noindent\textbf{Historical query collection form.}
We explore different methods to obtain historical object queries $\mathbf{O}^{his}_{t-k}$ in Table \ref{tab:connection-form}. The Recurrent method only utilizes historical queries from the last frame, which is described in Equation \ref{eq:hist-queries}; the Fully-Connected method collects all the historical object queries $\{\mathbf{\Bar{O}}_{t-N},...,\mathbf{\Bar{O}}_{t-1}\}$ but only uses them for the current frame $t$; the Dense method extends the Fully-Connected method with using all historical queries on every timestamp. These three variants can bring consistent improvement, proving the insensitivity of the collection form; the recurrent form obtains the best result, where mAP increased by 1.2\% and NDS increased by 1.3\%. The results also prove our hypothesis that historical queries from the previous frame provide a more reliable initialization.

% \noindent\textbf{Fusion start time.}
% Table \ref{tab:fusion-timestamp} explores the effect of when to start utilizing historical object queries. Note that if we want to begin using historical queries at timestamp $t-k$, we must collect historical queries from timestamp $t-k-1$. Given that we use 4 history frames in the ablation, the minimal start time is $t-3$. We observe that incorporating historical queries earlier could bring consistent gains, so we choose to fuse queries at the beginning in other experiments by default.
% % consistent with more history frames. still improve on the basis of merging history bev feature

% \begin{figure*}[thbp]
% % \begin{figure}[thbp]
% \centering
% \includegraphics[width=0.9 \textwidth]{HoP_arxiv/fig/vis_misplaced_2.pdf}
% \caption{Illustration of the temporal multi-stage decoder. Object queries are enhanced by their counterparts derived from the last frame, forming a recurrent connection. "TE BEV" denotes the temporal enhanced BEV feature.} 
% \label{Fig.vis-misplaced-} 
% % \end{figure}
% \end{figure*}

\subsection{Visualization}
Figure~\ref{Fig.vis-baseline} shows the detection result of the baseline method BEVFormer-opt and our proposed HoP-BEVFormer. The predictions are marked in yellow, while the ground truth boxes are marked in blue in the LIDAR\_TOP figure. We observe that HoP-BEVFormer successfully detects small objects or occluded objects thanks to the superior ability to aggregate temporal information. 

We also visualize the predictions of the object decoder with three variants of temporal decoder $\mathcal{T}$: only long-term, only short-term, and both of them in Figure~\ref{Fig.vis-ls}. 
The decoder with only a short-term branch can detect accurate attributes of objects, \eg, height and length, but fails to localize them precisely. 
Besides, results in the second row show that the short-term branch contributes to detecting more foreground objects because of its detailed spatial semantics. 
On the contrary, the decoder with only long-term modeling performs precise localization due to the long-term motion while struggling to detect the size of the object, \eg, height. 
When both long-term and short-term branches are adopted, we achieve the best result by combining their advantages.

%% file: sota.tex
\begin{table*}[t]
    \centering
    % \small
    \addtolength{\tabcolsep}{-3.0pt}
    \renewcommand{\arraystretch}{1.2}
    \begin{tabular}{l|c|c|cc|ccccc}
    \Xhline{1.0pt}
    Method & Backbone & TTA & NDS$\uparrow$ & mAP$\uparrow$ & mATE$\downarrow$ & mASE$\downarrow$ & mAOE$\downarrow$ & mAVE$\downarrow$ & mAAE$\downarrow$ \\
    \hline
    SOLOFusion~\cite{solofusion} & ConvNeXt-B~\cite{convnext} &  & 0.619 & 0.540 & 0.453 & 0.257 & 0.376 & 0.276 & 0.148 \\
    BEVFormerV2~\cite{bevformerv2} & InternImage-XL~\cite{internimage} &  & 0.648 & 0.580 & 0.448 & 0.262 & \textbf{0.342} & 0.238 & 0.128 \\
    BEVDet-Gamma~\cite{bevdet4d} & Swin-B~\cite{swin} & \cmark & 0.664 & 0.586 & 0.375 & \textbf{0.243} & 0.377 & 0.174 & \textbf{0.123} \\
    \rowcolor[gray]{.9}
    HoP-BEVDet4D-Depth & ViT-L~\cite{vit} &  & \textbf{0.685} &\textbf{0.624} & \textbf{0.367} &0.249 &0.354 &\textbf{0.171} &0.131 \\
    \Xhline{1.0pt}
    \end{tabular}
    \vspace{-1mm}
    \caption{Comparisons on the nuScenes camera-only 3D detection leaderboard. TTA denotes test-time augmentation.}
    \label{tab:app-optimizer}
    \normalsize
    \vspace{-1mm}
\end{table*}

%% file: component.tex
\begin{table*}[t]
    \centering\setlength{\tabcolsep}{6pt}
    \footnotesize
    \renewcommand{\arraystretch}{1.2}
    \resizebox{0.95\linewidth}{!}
    {
        \begin{tabular}{l|cc|ccccccc}
        \shline
        Method & HoP & HQ Fusion & NDS$\uparrow$ & mAP$\uparrow$ & mATE$\downarrow$ & mASE$\downarrow$ & mAOE$\downarrow$ & mAVE$\downarrow$ & mAAE$\downarrow$ \\
        \hline
        BEVFormer-opt &  &  & 0.513 & 0.414 & 0.655 & 0.274 & 0.447 & 0.368 & 0.194 \\ 
        BEVFormer-opt &  & \cmark & 0.525 & 0.425 & 0.660 & 0.274 & 0.395 & 0.357 & 0.188 \\ 
        BEVFormer-opt & \cmark &  & 0.531 & 0.428 & 0.638 & 0.269 & 0.352 & 0.369 & 0.185 \\ 
        BEVFormer-opt$^{\dagger}$ & \cmark &  & 0.539 & 0.435 & 0.629 & 0.268 & \textbf{0.342} & 0.360 & \textbf{0.184} \\
        \rowcolor[gray]{.9}
        BEVFormer-opt$^{\dagger}$ & \cmark & \cmark & \textbf{0.544} & \textbf{0.439} & \textbf{0.607} & \textbf{0.265} & 0.354 & \textbf{0.340} & 0.193 \\ 
        \hline
        % BEVDet4D &  &  &  & 0.324 & 0.687 & 0.293 & 0.712 & 0.338 & 0.220 \\ 
        % \rowcolor[gray]{.9}
        % BEVDet4D & \cmark &  & \textbf{0.484} & \textbf{0.376} & \textbf{0.655} & \textbf{0.278} & \textbf{0.623} & \textbf{0.283} & \textbf{0.203} \\ 
        % \hline
        BEVDet4D-Depth &  &  & 0.493 & 0.385 & 0.632 & 0.283 & 0.581 & 0.289 & 0.212 \\ 
        \rowcolor[gray]{.9}
        BEVDet4D-Depth & \cmark &  & \textbf{0.509} & \textbf{0.399} & \textbf{0.608} & \textbf{0.272} & \textbf{0.541} & \textbf{0.281} & \textbf{0.205} \\ 
        \shline
        \end{tabular}
    }
    \vspace{-2mm}
    \caption{\small{Component-wise ablations. $\dagger$ indicates we do not detach historical BEV features of last two layers in the BEV encoder. ``HQ Fusion'' refers to historical temporal query fusion.}
    }
    \label{tab:component}
    \vspace{-3mm}
\end{table*}

%% file: new_baseline.tex
\begin{table}[t]
    \centering\setlength{\tabcolsep}{6pt}
    \footnotesize
    \renewcommand{\arraystretch}{1.2}
    \resizebox{1\linewidth}{!}
    {
        \begin{tabular}{l|cc|cc}
        \shline
        \multirow{2}*{Method} & \multirow{2}*{NDS} & \multirow{2}*{mAP} & \#params & Training time \\
         & & & (M) & (sec/iter) \\
        \hline
        BEVFormer-R101-DCN & 0.517 & 0.416 & 78.6 & 2.7 \\ 
        \hline
        + Dropout 0 & 0.529 & 0.424 & 78.6 & 2.7 \\ 
        + Look forward twice & 0.531 & 0.430 & 78.6 & 2.7 \\ 
        + Smaller backbone R50 & 0.487 & 0.388 & 57.7 & 2.0 \\ 
        + Step LR decay & 0.493 & 0.391 & 57.7 & 2.0 \\ 
        + Data augmentation & 0.499 & 0.404 & 57.7 & 2.0 \\ 
        + Decoupled BEV encoder & 0.521 & 0.420 & 55.5  & 2.2 \\
        + Smaller resolution & 0.513 & 0.414 & 55.5 & 1.6 \\
        \hline
        BEVFormer-opt-R50 & \bf{0.513} & \bf{0.414} & \bf{55.5} & \bf{1.6} \\
        \shline
        \end{tabular}
    }
    \vspace{-2mm}
    \caption{\small{Step-by-step optimizing the BEVFormer baseline.}
    }
    \label{tab:tricks}
    \vspace{-5mm}
\end{table}

%% file: ablation.tex
% overall table of all ablations
\begin{table*}[t]
    \vspace{-.2em}
    \centering
    %#################################################
    % MAE decoder depth
    %#################################################
    \subfloat[
    \textbf{Temporal decoder design}. Both short-term and long-term can improve accuracy.
    \label{tab:temporal_decoder}
    ]{
    \centering
    \begin{minipage}{0.29\linewidth}
        \begin{center}
        \tablestyle{4pt}{1.05}
        \begin{tabular}{ccc}
        Case & NDS & mAP \\
        \shline
        None & 0.513 & 0.414  \\
        Short-term & 0.528 & 0.426 \\
        Long-term & 0.526 & 0.425 \\
        Both & \baseline{\textbf{0.531}} & \baseline{\textbf{0.428}} \\
        \end{tabular}
        \end{center}
    \end{minipage}
    }
    \hspace{2em}
    \subfloat[
    \textbf{Object decoder design}. The decoder can be flexibly implemented.
    \label{tab:obj_decoder}
    ]{
    \begin{minipage}{0.29\linewidth}{
        \begin{center}
        \tablestyle{4pt}{1.05}
        \begin{tabular}{ccc}
        Case & NDS & mAP \\
        \shline
        None & 0.513 & 0.414 \\
        Transformer & 0.521 & 0.421 \\
        ATSS & 0.529 & 0.426 \\
        CenterPoint & \baseline{\textbf{0.531}} & \baseline{\textbf{0.428}} \\
        \end{tabular}
        \end{center}}
    \end{minipage}
    }
    \hspace{2em}
    %#################################################
    % MAE with mask token on encoder
    %#################################################
    \subfloat[
    \textbf{Prediction target}.
    The explicit supervisions of 3D targets are critical.
    \label{tab:pred_target}
    ]{
    \begin{minipage}{0.29\linewidth}{
        \begin{center}
        \tablestyle{4pt}{1.05}
        \begin{tabular}{ccc}
        Case & NDS & mAP \\
        \shline
        None & 0.513 & 0.414 \\
        BEV feature & 0.518 & 0.418  \\
        3D objects & \baseline{\textbf{0.531}} & \baseline{\textbf{0.428}} \\
        \multicolumn{3}{c}{~}\\
        \end{tabular}
        \end{center}}
    \end{minipage}
    }
    \\
    \centering
    \vspace{.3em}
    %#################################################
    % MAE targets
    %#################################################
    \subfloat[
    \textbf{Prediction index}. Predicting the objects in frame $t$-$1$ achieves the best trade-offs.
    \label{tab:trunc_index}
    ]{
    \begin{minipage}{0.29\linewidth}{
        \begin{center}
        \tablestyle{6pt}{1.05}
        \begin{tabular}{cccc}
        $k$ & NDS & mAP & Memory \\
        \shline
        None & 0.513 & 0.414 & 1$\times$ \\
        -1 & 0.523 & 0.420 & 1.6$\times$ \\
        1 & \baseline{0.531} & \baseline{0.428} & \baseline{1.3$\times$} \\
        2 & 0.534 & 0.430 & \textbf{2.3}$\times$\\
        3 & \textbf{0.536} & \textbf{0.432} & \textbf{2.3}$\times$ \\
        \end{tabular}
        \end{center}}
    \end{minipage}
    }
    \hspace{2em}
    %#################################################
    % MAE data aug
    %#################################################
    \subfloat[
    \textbf{Training speed}. HoP slightly increases the training time.
    \label{tab:cost}
    ]{
    \centering
    \begin{minipage}{0.29\linewidth}
        {
        \begin{center}
        \tablestyle{4pt}{1.05}
        \begin{tabular}{ccc}
        % \Xhline{1.0pt}
        Case & NDS & Speed \\ \shline
        BEVFormer-opt & 0.513 & \textbf{1.6} \\
        HoP-BEVFormer-opt & \baseline{0.531} & \baseline{1.9} \\
        HoP-BEVFormer-opt$^{\dagger}$ & \textbf{0.539} & 2.1 \\
        BEVDet4D-Depth & 0.493 & 4.5 \\
        HoP-BEVDet4D-Depth & \baseline{0.509} & \baseline{4.7} \\
        % \Xhline{1.0pt}
        \end{tabular}
        \end{center}
        }
    \end{minipage}
    }
    \hspace{2em}
    %#################################################
    % MAE with mask types
    %#################################################
    \subfloat[
    \textbf{Historical query collection form}. All three connection forms bring improvements.
    \label{tab:connection-form}
    ]{
    \centering
    \begin{minipage}{0.29\linewidth}
        {
        \begin{center}
        \tablestyle{4pt}{1.05}
        \begin{tabular}{ccc}
        % \Xhline{1.0pt}
        Connection form & NDS  & mAP   \\ \shline
        None               & 0.513 & 0.414 \\
        Recurrent       & \baseline{\textbf{0.525}} & \baseline{\textbf{0.425}} \\
        Fully-Connected         & 0.521 & 0.422 \\
        Dense           & 0.523 & 0.421 \\
        \multicolumn{1}{c}{~}\\
        % \Xhline{1.0pt}
        \end{tabular}
        \end{center}
        }
    \end{minipage}
    }
    %#################################################
    \vspace{-.1em}
    \caption{\textbf{Ablation experiments} with ResNet-50 on nuScenes \textit{val}. Default settings are marked in \colorbox{baselinecolor}{gray}. $\dagger$ indicates we do not detach historical BEV features of last two layers in the BEV encoder.}
    \label{tab:ablations} \vspace{-.5em}
\end{table*}

% Fusion start time
    % \textbf{Fusion start time}. Fusing queries from the beginning works the best.
    % \label{tab:fusion-timestamp}
    % ]{
    % \begin{minipage}{0.29\linewidth}{
    %     \begin{center}
    %     \tablestyle{4pt}{1.05}
    %     \begin{tabular}{cccccc}
    %     % \Xhline{1.0pt}
    %     Start time & NDS & mAP \\ \shline
    %     None& 0.513 & 0.414 \\
    %     $t$   & 0.523 & 0.423 \\
    %     $t$-1 & 0.523 & 0.424 \\
    %     $t$-2 & 0.524 & 0.426 \\
    %     $t$-3 & \baseline{\textbf{0.527}} & \baseline{\textbf{0.426}} \\

    %     \end{tabular}
    %     \end{center}
    % }
    % \end{minipage}

%% file: 5_conclusions.tex
\section{Conclusion}
In this paper, we propose a new paradigm, named Historical Object Prediction (HoP) for multi-view 3D detection to leverage temporal information more effectively. 
The HoP approach is straightforward: given the current timestamp $t$, we generate a pseudo BEV feature of timestamp $t$-$k$ from its adjacent frames and utilize this feature to predict the object set at timestamp $t$-$k$. 
First, we elaborately design short-term and long-term temporal decoders, which can generate the pseudo BEV feature for timestamp $t$-$k$ without the involvement of its corresponding camera images. 
Second, an additional object decoder is flexibly attached to predict the object targets using the generated pseudo BEV feature. 
As a plug-and-play approach, HoP can be easily incorporated into state-of-the-art BEV detection frameworks, including BEVFormer and BEVDet series. 
Extensive experiments are conducted to evaluate the effectiveness of the proposed HoP on the nuScenes dataset.
Surprisingly, HoP achieves 68.5\% NDS and 62.4\% mAP on nuScenes test, outperforming all the 3D object detectors on the leaderboard by a significant margin. 